\documentclass{article}
\usepackage{ijcai20}

\usepackage{times}
\usepackage{soul}
\usepackage{url}
\usepackage[hidelinks]{hyperref}
\usepackage[utf8]{inputenc}
\usepackage[small]{caption}
\usepackage{graphicx}
\usepackage{amsmath}
\usepackage{amsthm}
\usepackage{booktabs}
\usepackage{algorithm}
\usepackage{algorithmic}
\usepackage{amssymb}
\usepackage{color,xcolor}
\usepackage{multirow}
\usepackage{array}
\title{A Similarity Inference Metric for RGB-Infrared Cross-Modality Person Re-identification}

\author{
Mengxi Jia$^1$
\and
Yunpeng Zhai$^1$\and
Shijian Lu$^2$\and
Siwei Ma$^{3,4}$\and
Jian Zhang$^{1,4}$\footnote{Corresponding author.}
\affiliations
$^1$School of Electronic and Computer Engineering, Peking University, China\\
$^2$Nanyang Technological University, Singapore\\
$^3$School of Electronics Engineering and Computer Science, Peking University, China\\
$^4$ Peng Cheng Laboratory, China\\
\emails
\{mxjia, ypzhai, swma, zhangjian.sz\}@pku.edu.cn,
shijian.lu@ntu.edu.sg
}

\begin{document}

\maketitle

\begin{abstract}
RGB-Infrared (IR) cross-modality person re-identification (re-ID), which aims to search an IR image in RGB gallery or vice versa, is a challenging task due to the large discrepancy between IR and RGB modalities. Existing methods address this challenge typically by aligning feature distributions or image styles across modalities, whereas the very useful similarities among gallery samples of the same modality (i.e. intra-modality sample similarities) are largely neglected. This paper presents a novel similarity inference metric (SIM) that exploits the intra-modality sample similarities to circumvent the cross-modality discrepancy targeting optimal cross-modality image matching. SIM works by successive similarity graph reasoning and mutual nearest-neighbor reasoning that mine cross-modality sample similarities by leveraging intra-modality sample similarities from two different perspectives. Extensive experiments over two cross-modality re-ID datasets (SYSU-MM01 and RegDB) show that SIM achieves significant accuracy improvement but with little extra training as compared with the state-of-the-art.
\end{abstract}

\section{Introduction}

\begin{figure}[t]
\begin{center}
  \includegraphics[width=1.0\linewidth]{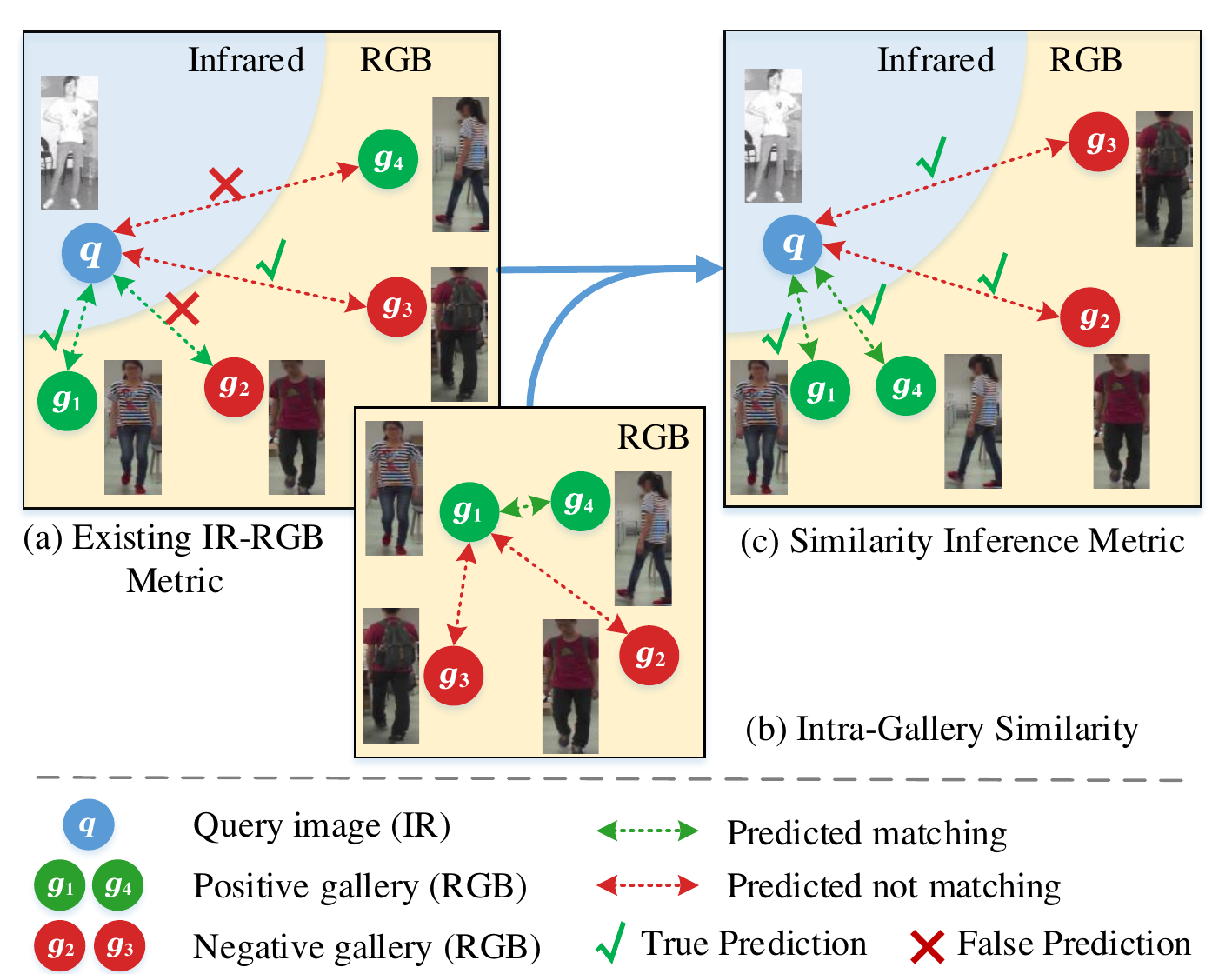}
\end{center}
\caption{Similarity Inference Metric (SIM) infers cross-modality sample similarities by exploiting intra-modality sample similarities. It enhances the existing IR-RGB metric to match the hard positive samples (\textit{e.g.} $g_4$) which are dissimilar from the query but similar to the predicted matching samples of the query (\textit{e.g. $g_1$}). }
\label{fig:motivation}
\end{figure}

Person re-identification (re-ID) is an important task in video surveillance. Given a query image of a person, re-ID aims to match persons
in an image gallery collected from non-overlapping camera networks.

To leverage the unique features of sensors of different modalities, cross-modality re-ID has been attracting increasing interest in recent years for more robust identification, e.g. by using infrared images as query and RGB images as gallery. On the other hand, cross-modality re-ID remains a challenging task due to the large discrepancy across image modalities in terms of distinct illumination, heterogeneous features, etc.

Two typical approaches have been explored to address the cross-modality re-ID challenges. The first approach attempts to align feature distribution of images of different modalities to reduce the cross-modality discrepancy~\cite{wu2017rgb}~\cite{ye2018hierarchical}~\cite{ijcai2018-152}~\cite{dai2018cross}. The other approach utilizes generative adversarial network (GAN) as a modality transformer to convert person images from one modality to another while preserving the identity information as much as possible ~\cite{wang2019learning}\cite{Wang_2020_AAAI}\cite{Wang_2019_ICCV}. These two types of approaches thus focus more on the reduction of cross-modality discrepancy or learning ID-preservative mapping across modalities, whereas the discriminative similarity among gallery samples of the same modality is largely neglected. The lack of this very useful information has become a major reason for the low performance of cross-modality person re-ID. 

In this paper, we propose an innovative similarity inference metric (SIM) for cross-modality person re-ID. SIM aims to infer cross-modality sample similarities by exploiting reliable intra-modality sample similarities as illustrated in Fig.~\ref{fig:motivation}. Instead of using the query-gallery similarity matrix for person matching like most existing methods do, we introduce similarity graph reasoning (SGR) and mutual nearest-neighbor reasoning (MNNR) that discover intra-modality sample similarities and circumvent the cross-modality discrepancy successively. Specifically, these two types of reasoning utilize the intra-modality similarities, in terms of graph shortest path and nearest neighbor overlap, to empower re-ID to match the hard positive samples which are dissimilar from the query but similar to the predicted matching samples of the query. What's more, SIM improves the cross-modality re-ID performance significantly and consistently. More details will be provided in the experimental section.

The main contributions of this work can be summarized in three aspects. \textit{First}, it proposes a similarity inference metric that successively improves cross-modality similarities by utilizing the discriminative intra-modality sample similarities. \textit{Second}, it designs novel similarity graph reasoning and mutual nearest-neighbor reasoning that can be applied to different cross-modality person re-ID metric with little extra training. \textit{Third}, it achieves significant performance improvement over the state-of-the-arts on two widely used cross-modality re-ID datasets: SYSU-MM01 and RegDB.

\section{Related Works}

\subsection{Single-modality Person Re-ID.}

Conventional re-ID research mainly focuses on the challenge of appearance variations in a single RGB modality, including illumination conditions, viewpoint variations, misalignment, etc. Existing methods can be broadly classified into two categories. Methods in the first category attempt to learn similarity metrics which are used to predict whether two images contain the same person~\cite{Zheng2011Person}~\cite{Zhen2013Learning}~\cite{Gou2014Person}~\cite{liao2015person}~\cite{chen2017beyond}\cite{hermans2017defense}. Methods in the second category focus on learning a discriminative feature representation, upon which efficient L$2$ or cosine distances can be applied~\cite{liao2015person}~\cite{zhao2017spindle}~\cite{li2018harmonious}~\cite{zhai_adcluster}~\cite{DBLP:journals/pr/YangYLJXG19}. Besides, most existing methods were developed for single-modality re-ID which cannot tackle the cross-modality re-ID well due to the large discrepancy across modalities.

\subsection{Cross-modality Person Re-ID.}

For the RGB-Infrared cross-modality re-ID, the discrepancies come not just from appearance variations but also from heterogeneous images of different modalities. Two typical approaches have been explored to reduce the cross-modality discrepancies. The first approach attempts to align the feature distribution of images of different modalities. For example, \cite{wu2017rgb} explores three different network structures and uses deep zero-padding for evolving domain-specific nodes. \cite{ye2018hierarchical} jointly optimizes the modality-specific and modality-shared metrics to learn multi-modality representations. \cite{ijcai2018-152} proposes a dual-path network with a bi-directional dual-constrained top-ranking loss to learn common representations. \cite{dai2018cross} designs a cross-modality generative adversarial network (cmGAN) to learn discriminative representations from different modalities. \cite{hao2019hsme} proposes a hyper-sphere manifold embedding model. The second approach instead uses generative adversarial network (GAN) as a modality transformer to convert person images from one modality to another while preserving the identity information as much as possible~\cite{wang2019learning}~\cite{Wang_2019_ICCV}~\cite{Wang_2020_AAAI}.

Though these methods reduce the modality discrepancies, the very useful discriminative similarity among gallery samples of the same modality is largely neglected. Our similarity inference metric captures such intra-gallery sample similarity which improves the cross-modality re-ID significantly, more details to be discussed in the ensuing subsections.

\subsection{Re-ranking for Person Re-ID.}
Re-ranking methods have been wildly studied to improve conventional person re-ID. After an initial ranking list is obtained, re-ranking aims to raise the rank of relevant images in an automatic and unsupervised manner. Recently, various re-ranking methods have been reported. For example, \cite{garcia2015person} learns an unsupervised re-ranking model that removes the visual ambiguities by analyzing the content and context information in the initial ranking list. \cite{ye2016person} attempts to tackle the re-ranking problem by exploiting the common nearest neighbors. To address the false match issue from $k$-nearest neighbors, \cite{Zhong_2017_CVPR} proposes to utilize $k$-reciprocal neighbors and designs an encoding method to revise the initial rank list by calculating feature distance and jaccard distance of samples. 

Most existing re-ranking methods are designed for single-modality re-ID which do not work well in the cross-modality re-ID task. The major problem is that existing re-ranking methods cannot re-rank the samples of different modalities which have different similarity metrics as compared with samples of a single modality. We tackle this problem by combining cross-modality $k$-nearest neighbors and intra-modality $k$-reciprocal neighbors which improves the re-ID performance significantly, more details to be described in Sec.\ref{sec:mnnr}.

\section{The Proposed Approach}

\begin{figure*}[t]
\begin{center}
  \includegraphics[width=1.0\linewidth]{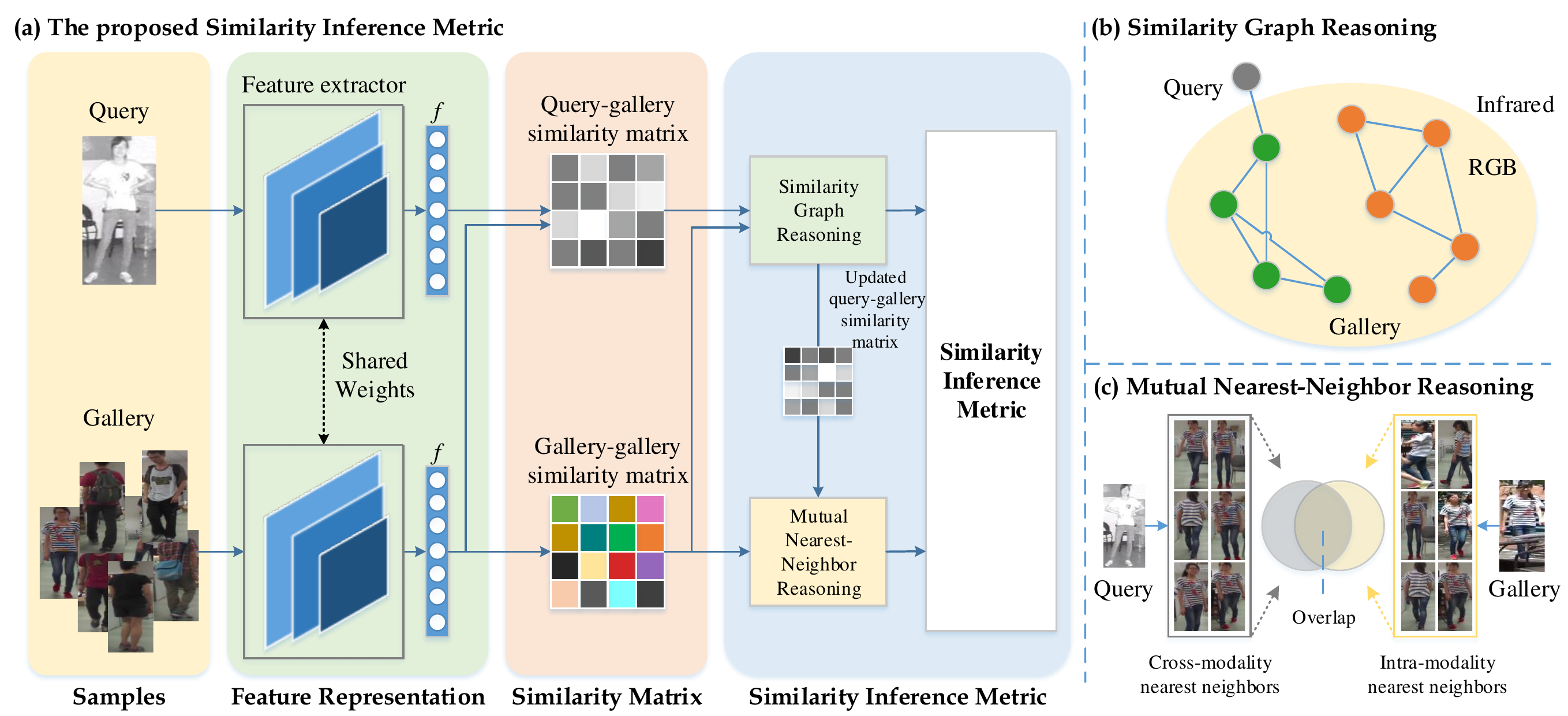}
\end{center}
\caption{(a) The flowchart of the proposed similarity inference metric (SIM): SIM consists of two types of reasoning including similarity graph reasoning and mutual nearest-neighbor reasoning. (b) Similarity graph reasoning introduced in Sec.~\ref{sec:sgr}, where each color (green or orange) refers to samples of the same ID. (c) Mutual nearest-neighbor reasoning introduced in Sec.~\ref{sec:mnnr}.} 
\label{fig:flowchart}
\end{figure*}
Given a query infrared person image $\boldsymbol{q}$ and the gallery set with $N_g$ RGB images $\boldsymbol{\mathcal{G}} = \{\boldsymbol{g}_j~|~j =1,2,..., N_g\} $, cross-modality re-ID ranks the gallery images according to their similarities to the query $\boldsymbol{q}$. Existing methods usually derive the similarity metric by directly comparing features of cross-modality samples, which often face low precision due to the gap and bias across image modalities. The proposed Similarity Inference Metric (SIM) aims to improve the cross-modality similarity metrics by exploiting the discriminative intra-modality similarity among gallery samples. It consists of feature representation, similarity graph reasoning, and mutual nearest-neighbor reasoning as illustrated in Fig. \ref{fig:flowchart}, more details to be described in the ensuing subsections.

\subsection{Feature Representation}

A weight-sharing two-stream CNN structure is designed to learn features and image representation from infrared and RGB images as illustrated in Fig. \ref{fig:flowchart}a. The CNN models are trained by optimizing cross entropy loss and triplet loss with infrared and RGB training samples together. 


In inference phase, each infrared query image $\boldsymbol{q}_i$ in the query set $\boldsymbol{\mathcal{Q}} = \{\boldsymbol{q}_i~|~i=1,2,...,N_q \}$ and each RGB gallery image $\boldsymbol{g}_j$ from $\boldsymbol{G}$ are fed to the trained model to extract respective features $\boldsymbol{f}_{\boldsymbol{q}_i}$ and $\boldsymbol{f}_{\boldsymbol{g}_j}$. A query-gallery similarity matrix $\boldsymbol{D}_{\boldsymbol{q},\boldsymbol{g}}\in \mathbb R^{N_q\times N_g}$ can then be obtained by computing $L_2$ distance between all query images and gallery images, where each matrix element $\boldsymbol{D}(i,j)$ denotes the distance between $\boldsymbol{f}_{\boldsymbol{q}_i}$ and $\boldsymbol{f}_{\boldsymbol{g}_j}$. Similarly, a gallery-gallery similarity matrix $\boldsymbol{D}_{\boldsymbol{g},\boldsymbol{g}}\in \mathbb R ^{N_g\times N_g}$ can be obtained for all image pairs in the gallery set. 
Due to abundant optical information with little modality gap,  $\boldsymbol{D}_{\boldsymbol{g},\boldsymbol{g}}$ is much more discriminative than $\boldsymbol{D}_{\boldsymbol{q},\boldsymbol{g}}$ as intuitively shown in Table~\ref{table:gap}.



\begin{table}[t]
\begin{center}
\begin{tabular}{c|cc}
    \hline\hline
    Settings  & mAP & Rank-1  \\
    \hline
    Cross-modality ($D_{q,g}$)       &38.78 & 54.67  \\
    Internal-modality ($D_{g,g}$)      &88.84 & 97.67    \\
    \hline\hline
\end{tabular}
\end{center}
\vspace{-0.4cm}
\caption{We evaluate mAP and rank-1 on SYSU-MM01 dataset using infrared and RGB images as query respectively. Results demonstrate better performance of intra-modality similarities.}
\label{table:gap}
\vspace{-0.4cm}
\end{table}

\subsection{Similarity Graph Reasoning}
\label{sec:sgr}

We propose similarity graph reasoning to circumvent the cross-modality discrepancy by leveraging the intra-modality similarity. The idea is that for a query image $\boldsymbol{q}$ and its similar gallery image $\boldsymbol{g}$, other gallery images that are similar to $\boldsymbol{g}$ should be similar to $\boldsymbol{q}$ (via $\boldsymbol{g}$) even though they may have large distances from $\boldsymbol{q}$ as illustrated in Fig.~\ref{fig:flowchart}b. 
With the matrices $\boldsymbol{D}_{\boldsymbol{q},\boldsymbol{g}}$ and $\boldsymbol{D}_{\boldsymbol{g},\boldsymbol{g}}$ as described in the previous subsection, we define a similarity graph $\mathbf{A}(\boldsymbol{\mathcal{V}}, \boldsymbol{\mathcal{E}})$ on the whole test set including all query and gallery, where each node in $\boldsymbol{\mathcal{V}}=\{\boldsymbol{\mathcal{Q}};\boldsymbol{\mathcal{G}}\}$ represents an image sample  and each edge in $\boldsymbol{\mathcal{E}}$ represents the similarity between its connected two nodes. We initialize the cross-modality edges (query-gallery) with $\boldsymbol{D}_{\boldsymbol{q},\boldsymbol{g}}$ and intra-modality edges (gallery-gallery) with $\boldsymbol{D}_{\boldsymbol{g},\boldsymbol{g}}$ as follows:

\begin {equation} 
    \left\{ 
    \begin{aligned}
    \boldsymbol{\mathcal{E}}(\boldsymbol{q}_i, \boldsymbol{g}_j) &= \boldsymbol{D}_{\boldsymbol{q},\boldsymbol{g}}(\boldsymbol{q}_i, \boldsymbol{g}_j)\\
     \boldsymbol{\mathcal{E}}(\boldsymbol{g}_j,\boldsymbol{g}_k) &= \lambda \boldsymbol{D}_{\boldsymbol{g},\boldsymbol{g}}(\boldsymbol{g}_j, \boldsymbol{g}_k),
     \end{aligned}
     \right.
\label{eq:graph}
\end {equation}
where $\lambda \in [0, 1]$ is a scale factor that adjusts the ratio of two distance space. 

Given the query image $\boldsymbol{q}_i$ and the gallery image $\boldsymbol{g}_j$, the distance $d(\boldsymbol{q}_i, \boldsymbol{g}_j)$ in perspective of similarity graph reasoning is defined as the shortest path from $\boldsymbol{q}_i$ to $\boldsymbol{g}_j$ between the query node $\boldsymbol{q}_i$ and the gallery nodes $\boldsymbol{\mathcal{G}}$ on the Graph $\mathbf{A}$. To be specific, suppose $\boldsymbol{\Omega}_{\boldsymbol{q}_i,\boldsymbol{g}_j}$ denotes the set that includes all the possible path from $\boldsymbol{q}_i$ to $\boldsymbol{g}_j$. For any path $\mathcal{P} \in \boldsymbol{\Omega}_{\boldsymbol{q}_i,\boldsymbol{g}_j}$, $\mathcal{P}=(\boldsymbol{p}_1, \boldsymbol{p}_2, ..., \boldsymbol{p}_n)$ where $n=length(\mathcal{P}), \boldsymbol{p}_1 = \boldsymbol{q}_i, \boldsymbol{p}_n = \boldsymbol{g}_j, \boldsymbol{p}_k \in \boldsymbol{\mathcal{G}}~(2\leq k \leq n -1)$, $d(\boldsymbol{q}_i, \boldsymbol{g}_j)$ is formulated as 
\begin {equation} 
    \begin{aligned}
    d(\boldsymbol{q}_i, \boldsymbol{g}_j) = \mathop{\min}_{\mathcal{P} \in \boldsymbol{\Omega}_{\boldsymbol{q}_i,\boldsymbol{g}_j} } \sum_{t=1}^{n-1} \boldsymbol{\mathcal{E}}(\boldsymbol{p}_t, \boldsymbol{p}_{t+1}).
     \end{aligned}
\end {equation}

Due to the fact that $L_2$ metric used in gallery satisfies triangle inequality below
\begin {equation} 
    \begin{aligned}
     &\boldsymbol{\mathcal{E}}(\boldsymbol{g}_i, \boldsymbol{g}_j)+\boldsymbol{\mathcal{E}}(\boldsymbol{g}_j, \boldsymbol{g}_t)\geq \boldsymbol{\mathcal{E}}(\boldsymbol{g}_i, \boldsymbol{g}_t), &\forall ~1\leq i, j, t\leq N_g.
     \end{aligned}
\end {equation}

Thus, the query-gallery distance can be simplified by:
\begin {equation} 
    \begin{aligned}
    d(\boldsymbol{q}_i, \boldsymbol{g}_j) = \mathop{\min}_{1 \leq t \leq N_g} \{ \boldsymbol{\mathcal{E}}(\boldsymbol{q}_i, \boldsymbol{g}_t) + \boldsymbol{\mathcal{E}}(\boldsymbol{g}_t, \boldsymbol{g}_j) \}.
     \end{aligned}
\end {equation}

Further, we use the mean of the first $K$ shortest paths instead of the shortest one for more stable cross-modality distance evaluation as follows:
\begin {equation} 
    \begin{aligned}
    d_S(\boldsymbol{q}_i, \boldsymbol{g}_j) = \frac{1}{K} \sum_{k=1}^K d^{(k)}(\boldsymbol{q}_i, \boldsymbol{g}_j).
     \end{aligned}
\label{eq:sgr}
\end {equation}
where $d^{(k)}(\boldsymbol{q}_i, \boldsymbol{g}_j)$ denotes the $k$-th shortest path between $\boldsymbol{q}_i$ and $\boldsymbol{g}_j$. 

In practice, to reduce the computational complexity, all useless edges between gallery pairs are deleted except those between each sample and its $k$-nearest neighbors in the gallery set.  

\begin{algorithm}[tb]
\caption{Similarity Inference Metric (SIM)}
\label{alg:algorithm}
\textbf{Input}: Query-gallery similarity matrix $\boldsymbol{D}_{\boldsymbol{q},\boldsymbol{g}}$, gallery-gallery similarity matrix $\boldsymbol{D}_{\boldsymbol{g},\boldsymbol{g}}$\\
\textbf{Parameter}: $\lambda$, $K$ and  $\alpha$ \\
\textbf{Output}: SIM $\boldsymbol{d}_{SIM}$ 
\begin{algorithmic}[1] 
\STATE Initialize similarity graph $\mathbf{A}(\boldsymbol{\mathcal{V}} ,\boldsymbol{\mathcal{E}})$ as Eq.~(\ref{eq:graph}).
\STATE \textit{$\%$Compute SGR distance} 
\FOR{each $\boldsymbol{q}_i$, $\boldsymbol{g}_j$} 

\FOR{$\boldsymbol{g}_t$ in $\boldsymbol{g}_j$'s k-nearest neighbors}
\STATE Calculate $d(\boldsymbol{q}_i, \boldsymbol{g}_t, \boldsymbol{g}_j) = \boldsymbol{\mathcal{E}}(\boldsymbol{q}_i, \boldsymbol{g}_t)+\boldsymbol{\mathcal{E}}(\boldsymbol{g}_t, \boldsymbol{g}_j)$
\ENDFOR
\STATE Sort $d(\boldsymbol{q}_i, \boldsymbol{g}_t, \boldsymbol{g}_j)$ for all $\boldsymbol{g}_t$.
\STATE Calculate $\boldsymbol{d}_{S}(\boldsymbol{q}_i, \boldsymbol{g}_j)$ according to Eq.~(\ref{eq:sgr}).
\ENDFOR
\STATE \textit{$\%$Compute MNNR distance} 
\FOR{each $\boldsymbol{q}_i$, $\boldsymbol{g}_j$}
\STATE Calculate $\boldsymbol{d}_{M}(\boldsymbol{q}_i, \boldsymbol{g}_j)$ according to Eq.~(\ref{eq:mnnr}).
\ENDFOR
\STATE Calculate $\boldsymbol{d}_{SIM}$ according to Eq.~(\ref{eq:sim}).
\STATE \textbf{return} $\boldsymbol{d}_{SIM}$
\end{algorithmic}
\end{algorithm}

\subsection{Mutual Nearest-Neighbor Reasoning}
\label{sec:mnnr}
We proposed mutual nearest-neighbor reasoning (MNNR) under the hypothesis that a query image $\boldsymbol{q}_i$ and a gallery image $\boldsymbol{g}_j$ are more likely to be a true match if they have the same mutual $k$-nearest neighbors in the gallery set. Neighbor information has been explored in re-ranking based re-ID, e.g. by $k$-reciprocal encoding \cite{Zhong_2017_CVPR}. But it was mainly used for single-modality re-ID which does not work well in cross-modality re-ID where similarity metrics of query-gallery and gallery-gallery are discrepant. For example, the cross-modality query-gallery distance $D_{\boldsymbol{q},\boldsymbol{g}}$ and the intra-modality gallery-gallery distance $D_{\boldsymbol{g},\boldsymbol{g}}$ are often at different scales and cannot be ranked while handling test samples of different modalities.

MNNR employs a series of asymmetric strategies to handle the cross-modality discrepancy as shown in Fig.~\ref{fig:flowchart}c. First, it uses gallery set as the search space without including query. For an IR query $\boldsymbol{q}$, it ranks gallery images with similarities $d_{S}$ and obtains its $k_{\boldsymbol{q}}$ cross-modality nearest neighbors:
\begin {equation} 
    \begin{aligned}
    \boldsymbol{\mathcal{N}}_c(\boldsymbol{q}_i, k_{\boldsymbol{q}}, d_{S}) = \{\boldsymbol{g}^{(1)},\boldsymbol{g}^{(2)},...,\boldsymbol{g}^{(k_q)}\}. 
     \end{aligned}
\end {equation}
For a RGB gallery image $\boldsymbol{g}$, it ranks the gallery images with
$\boldsymbol{D}_{\boldsymbol{g},\boldsymbol{g}}$ and obtains its
$k_{\boldsymbol{g}}$ intra-modality reciprocal nearest neighbors as
$\boldsymbol{\mathcal{R}}^*_i(\boldsymbol{g}, k_{\boldsymbol{g}},  \boldsymbol{D}_{\boldsymbol{g},\boldsymbol{g}})$. 
The mutual nearest neighbors of $\boldsymbol{q}$ and $\boldsymbol{g}$ can thus be defined by the overlap between $\boldsymbol{\mathcal{N}}_c(\boldsymbol{q}, k_{\boldsymbol{q}}, d_{S})$ and $\boldsymbol{\mathcal{R}}^*_i(\boldsymbol{g}, k_{\boldsymbol{g}}, \boldsymbol{D}_{\boldsymbol{g},\boldsymbol{g}})$. Intuitively, more mutual nearest neighbors means higher similarity and the MNNR distance $d_{M}$ can be defined by:
\begin {equation} 
    \begin{aligned}
    d_{M}(\boldsymbol{q}_i, \boldsymbol{g}_j) = 1 - \frac{|\boldsymbol{{\mathcal{N}}_c}(\boldsymbol{q}_i, k_{\boldsymbol{q}}, d_{S}) \cap \boldsymbol{\mathcal{R}}^*_i(\boldsymbol{g}_j, k_{\boldsymbol{g}}, \boldsymbol{D}_{\boldsymbol{g},\boldsymbol{g}})|}{|\boldsymbol{{\mathcal{N}}_c}(\boldsymbol{q}_i, k_{\boldsymbol{q}}, d_{S}) \cup \boldsymbol{\mathcal{R}}^*_i(\boldsymbol{g}_j, k_{\boldsymbol{g}}, \boldsymbol{D}_{\boldsymbol{g},\boldsymbol{g}})|}.
     \end{aligned}
\label{eq:mnnr}
\end {equation}
where $|\cdot |$ denotes the number of candidates in the set.

\subsection{Similarity Inference Metric}
The proposed Similarity Inference Metric can thus be derived by combining the similarity graph reasoning and mutual nearest-neighbor reasoning. It jointly aggregates $d_{S}$ and $d_{M}$ as the final distance as follows:
\begin {equation} 
    \begin{aligned}
    \boldsymbol{d}_{SIM} = \alpha \boldsymbol{d}_{S} + (1 - \alpha) \boldsymbol{d}_{M}.
     \end{aligned}
\label{eq:sim}
\end {equation}
where $\alpha \in [0, 1]$ denotes the penalty factor. When $\alpha = 1$, only the similarity graph reasoning is considered. 
\textbf{Algorithm 1} provides the detailed description of our proposed similarity inference metric.

\subsection{Complexity Analysis}
Most of computations focus on pairwise distance calculation and distance ranking for all gallery pairs and the computation complexity is $O(N_g^2)$ and $O(N_g^2\log N_g)$, respectively. Given a new query $\boldsymbol{q}$, SIM just computes the distance between $\boldsymbol{q}$ and gallery ($O(N_g)$), ranks all path distance for SGR ($O(K\log K)$), computes the $k_{\boldsymbol{q}}$-nearest neighbors for MNNR ($O(N_g\log N_g)$), and ranks the final distance ($O(N_g \log N_g)$).

\section{Experiments}



\subsection{Datasets and Settings}
The proposed SIM is evaluated over two public datasets RegDB and SYSU-MM01. The  standard Cumulative Matching Characteristic (CMC) curve and mean average precision (mAP) are adopted as the evaluation metrics. Different from the traditional single-modality re-ID, the evaluations here use IR images as probe set and RGB images as gallery set for both datasets.

\textbf{RegDB} \cite{nguyen2017person} is collected by using dual cameras (with optical and thermal sensors). It contains images of 412 persons, where for each person 10 RGB images and 10 IR images are collected. Following the evaluation protocol in (Ye et al.2018), this dataset is randomly split into two halves, one half for training and the other half for testing. In addition, the evaluation procedure is repeated for 10 trials to achieve statistically stable results.

\textbf{SYSU-MM01} \cite{wu2017rgb} is a large-scale RGB-IR re-ID dataset which contains images of 419 identities captured using six disjoint surveillance cameras (four RGB cameras and two IR cameras). The training set contains 19,659 RGB images and 12,792 IR images of 395 persons and the test set contains images of 96 persons. Following \cite{wu2017rgb}, we adopt the multi-shot \textit{all-search} mode evaluation protocol where 10 images of a person are randomly selected to form the gallery set with 10 times repeat.

\textbf{Implementation details.}
We adopt the ResNet-50 \cite{he2016deep} as the backbone network and initialize it by using parameters pre-trained on the ImageNet \cite{krizhevsky2012imagenet}. During training, the input image is uniformly resized to $256 \times 128$ and traditional image augmentation is performed via random flipping and random erasing \cite{zhong2017random}. 
In addition, we use the Adam optimizer to train the model and the learning rate is set at $3.5\times 10^{-4}$. The whole training process consists of 200 epochs.

\subsection{Comparison with State-of-the-Arts}

\begin{table}[t]
\begin{center}
\begin{tabular}{l|p{1.1cm}<{\centering}p{1.1cm}<{\centering}|p{1.1cm}<{\centering}p{1.1cm}<{\centering}}
    \hline\hline
    \multirow{2}{*}{\textbf{Methods}} & \multicolumn{2}{c}{\textbf{Visible2Thermal}} & \multicolumn{2}{c}{\textbf{Thermal2Visible}} \\
    \cline{2-5}
    & mAP & Rank-1 & mAP & Rank-1 \\
    \hline\hline

    LOMO      & 2.28 & 0.85  & - & -  \\ 
    HOG       & 10.31 & 13.49  & - & -  \\ 
    GSM       & 15.06 & 17.28  & - & -\\ 
    \hline

    One-stream       & 14.02 & 13.11  & - & -   \\
    Two-stream       & 12.43 & 30.36  & - & -  \\
    Zero-Padding   & 18.90  & 17.75  & 17.82 & 16.63 \\
    TONE            & 14.92  & 16.87  & 16.98 & 13.86  \\
    HCML            & 20.08  & 24.44  & 22.24 & 21.70  \\
    BDTR            & 31.83  & 33.47  & 31.10 & 32.72 \\
    D-HSME         & 47.0  & 50.9  & 46.2 & 50.2  \\
    \hline

    D$^2$RL                  & 44.1 & 43.4  & 44.1 & 43.4 \\
    PIG             & 49.3  & 48.5  & 48.9 & 48.1 \\
    AlignGAN        & 53.60 & 57.90  & 53.40 & 56.30    \\
    \hline
    SIM (ours)    & \textbf{75.29} & \textbf{74.47}  & \textbf{78.30} & \textbf{75.24} \\ 
    \hline\hline
\end{tabular}
\end{center}
\vspace{-0.3cm}
\caption{Comparison with state-of-the-art cross-modality re-ID methods over the dataset RegDB: Visible2Thermal means using RGB images as query and IR images as gallery, and Thermal2Visible means the opposite.}
\label{table:sota_regdb} 
\vspace{-0.3cm}
\end{table}

The proposed SIM is compared with a number of cross-modality person re-ID methods that can be broadly classified into three categories: 1) LOMO~\cite{liao2015person}, HOG~\cite{dalal2005histograms} and GSM~\cite{lin2016cross} that use hand-crafted features; 2) One-stream, Two-stream, Zero-Padding~\cite{wu2017rgb}, TONE~\cite{ye2018hierarchical}, BDTR~\cite{ijcai2018-152} and cmGAN~\cite{dai2018cross} that focus on feature distribution alignment; and 3) D$^2$RL~\cite{wang2019learning}, PIG~\cite{Wang_2020_AAAI} and AlignGAN~\cite{Wang_2019_ICCV} that use GANs to transfer image styles. Table~\ref{table:sota_regdb} and Table~\ref{table:sota_sysu} show the experimental results over the datasets RegDB and SYSU-MM01, respectively, where Visible2Thermal means using RGB images as query and IR images as gallery, and Thermal2Visible means the opposite.

\begin{table}[t]
\begin{center}
\begin{tabular}{l|p{2cm}<{\centering}p{2cm}<{\centering}}
    \hline\hline
    \textbf{Methods} & mAP & Rank-1 \\
    \hline
 
    LOMO      & 2.28 & 4.70  \\ 
    HOG       & 2.16 & 3.82    \\ 
    GSM       & 4.38 & 6.19   \\ 
    \hline

    One-stream  & 8.59 & 16.3   \\
    Two-stream  & 8.03 & 16.4  \\
    Zero-Padding & 10.9 &19.2 \\
    cmGAN  & 22.27  & 31.49 \\
    \hline

    PIG       & 29.5  & 45.1  \\
    AlignGAN  & 33.90 & 51.50  \\
    \hline
    SIM (ours)    &\textbf{60.88} & \textbf{56.93} \\ 
    \hline\hline
\end{tabular}
\end{center}
\vspace{-0.4cm}
\caption{Comparison with state-of-the-art cross-modality re-ID methods over the dataset SYSU-MM01} 
\label{table:sota_sysu}
\end{table}

As the two tables show, methods in the first category do not perform well due to the constraints of hand-crafted features. Methods in the second category learn modality-invariance features by suppressing feature distribution gaps across modality, which achieve clearly better re-ID performance. GAN based methods reduce the modality discrepancy at image level which further improve the re-ID performance. 

Our proposed Similarity Inference Metric outperforms all the competing methods significantly. As Table~\ref{table:sota_regdb} shows, it outperforms the state-of-the-art (AlignGAN) by 24.9\% in mAP (78.3\% vs 53.4\%) and 18.94\% in rank-1 accuracy (75.24\% vs 56.3\%) for Thermal2Visible. Similar improvement is obtained for the Visible2Thermal. For the dataset SYSU-MM01, SIM obtains an mAP of 60.88\% and a rank-1 of 56.93\% as shown in Table~\ref{table:sota_sysu}, which outperforms the state-of-the-art (AlignGAN) by 26.98\% and 5.43\%, respectively.

\subsection{Ablation Studies}

\begin{table}[t]
\begin{center}
\begin{tabular}{l|cc|p{1.1cm}<{\centering}p{1.1cm}<{\centering}}
    \hline\hline
    Methods & SGR & MNNR & mAP & Rank-1 \\
    \hline
    Baseline     & $\times$ & $\times$ &39.90 & 53.93    \\ 
    \hline
    SGR only     & $\checkmark$ & $\times$ & 59.17 & 56.62   \\
    MNNR only     & $\times$ & $\checkmark$ & 54.39 & 56.31    \\
    SIM     & $\checkmark$ & $\checkmark$ & 60.88 & 56.93    \\ 
    \hline\hline
\end{tabular}
\end{center}
\vspace{-0.3cm}
\caption{Ablation studies of our proposed Similarity Inference Metric over SYSU-MM01: \textit{Baseline} uses the traditional metric, i.e. $L_2$ distance between image features; \textit{SGR only} incorporates the Similarity Graph Reasoning (SGR) only over the \textit{Baseline}; \textit{MNNR only} incorporates the Mutual Nearest-Neighbor Reasoning (MNNR) only over the \textit{Baseline}; \textit{SIM} incorporates both SGR and MNNR.}
\label{table:ablation}
\vspace{-0.4cm}
\end{table}

Extensive ablation studies have been performed to evaluate each component of our proposed SIM. As Table~\ref{table:ablation} shows, four networks are trained including: 1) \textit{Baseline} that uses the traditional L$_2$ distance to measure the feature similarity; 2) \textit{SGR only} that just incorporates the Similarity Graph Reasoning (as described in Section~\ref{sec:sgr}) beyond the \textit{Baseline}; 3) \textit{MNNR only} that just incorporates the proposed Mutual Nearest-Neighbor Reasoning (as described in Section~\ref{sec:mnnr}) beyond the \textit{Baseline}; and \textit{SIM} that incorporates both SGR and MNNR. As Table~\ref{table:ablation} shows, the \textit{Baseline} does not perform well due to the large discrepancy across image modalities. 

In addition, either \textit{SGR only} or \textit{MNNR only} improves the re-ID performance greatly. Specifically, \textit{SGR only} achieves a mAP of 59.17\% and a rank-1 accuracy of 56.62\%, which are higher than the \textit{Baseline} by 19.27\% and 2.69\%, respectively. This results show that SGR improves the sample similarity greatly by exploiting the discriminative within-gallery similarities. Similarly, \textit{MNNR only} improves the mAP by 14.49\% and the rank-1 accuracy by 2.37\%, respectively, as compared with the \textit{Baseline}. The effectiveness of the MNNR can be largely attributed to the use of the overlap of k-nearest neighbor sets in gallery between image pairs.

\begin{figure}[t]
\begin{center}
  \includegraphics[width=1.0\linewidth]{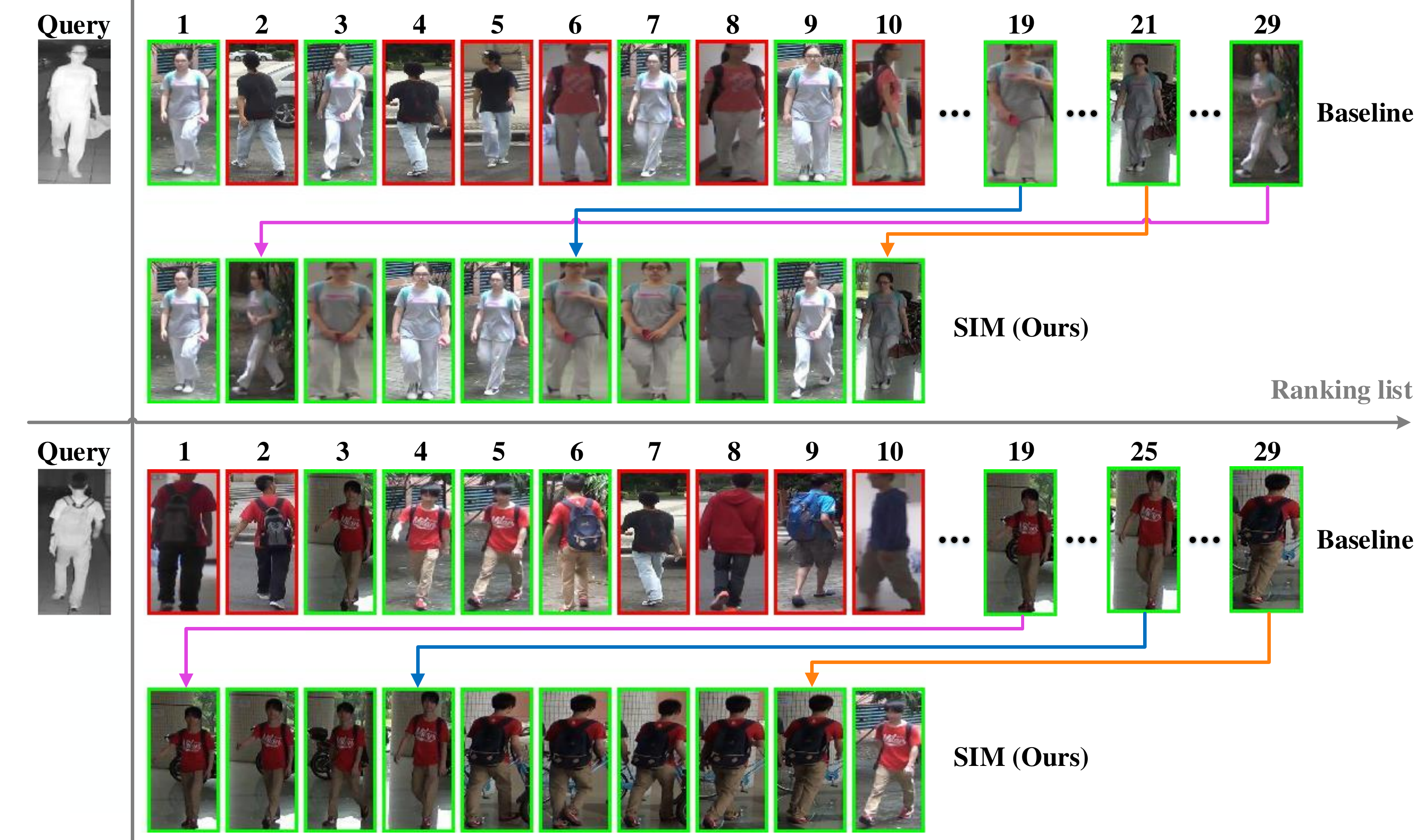}
\end{center}
\vspace{-3mm}
\caption{Illustration of how SIM helps improve cross-modality re-ID on the SYSU-MM01 dataset: With SIM, the similarity of IR and RGB images is improved greatly – the three lines in blue, orange, and pink show three examples of improved image similarities by our proposed SIM. Person images in red (green) boxes denote the negative (positive) samples.
} 
\label{fig:example}
\vspace{-4mm}
\end{figure}

Further, \text{SIM} with both SGR and MNNR outperforms either \textit{SGR only} or \textit{MNNR only}, demonstrating the complementariness of the two proposed reasoning mechanisms. It achieves a mAP of 60.88\% and a rank-1 accuracy of 56.93\% which are higher than the \textit{Baseline} by 20.98\% and 3.00\%, respectively. This shows that the proposed SIM enhances the cross-modality sample similarities effectively.
The contribution of our SIM can also be observed in the ranking list as illuminated in Fig.~\ref{fig:example}. SIM effectively improves the similarities of true persons which are ranked behind of the baseline.

\subsection{Parameters Analysis}

\begin{figure}[t]
\begin{center}
  \includegraphics[width=1.0\linewidth]{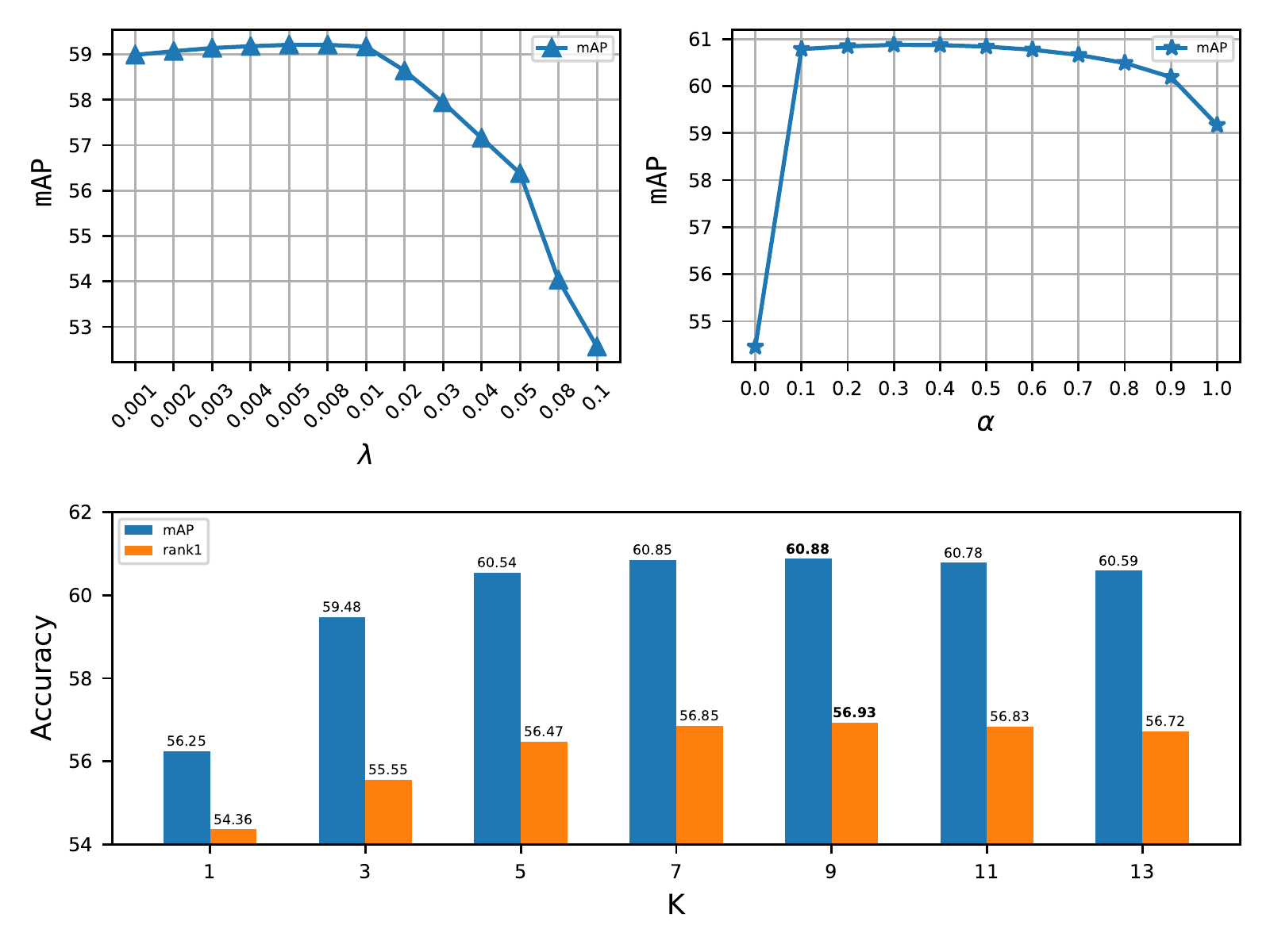}
\end{center}
\vspace{-6mm}
\caption{The impact of the parameter $\lambda$, $\alpha$ and $K$ on re-ID performance on the SYSU-MM01 dataset.} 
\label{fig:parameters}
\vspace{-3mm}
\end{figure}

The proposed SIM involves three key parameters including scale factor $\lambda$, limit factor $K$ and penalty factor $\alpha$. The three parameters are studied by setting them to different values and checking the corresponding re-ID performance as shown in Fig.~\ref{fig:parameters}. As the top-left graph shows, $\lambda$ should be small so as to exploit the within-gallery sample similarity sufficiently ($\alpha$ and $K$ fixed at $0.01$ and $9$). This can also be observed for $\alpha$ and $K$. For example, when $K$ is set at a small value 1, the false positive matching increases clearly as it lowers the fault tolerance for the first matching. On the contrary, the re-ID performance is degraded due to its weak discrimination when $K$ is set at 13. Experiments show that SIM performs optimally when $\lambda = 0.01$, $K = 9$ and $\alpha = 0.3$.






\subsection{Generalization Analysis}

The proposed SIM is a generic metric that can work with different existing re-ID methods. We study this nice property by applying SIM to AlignGAN \cite{Wang_2019_ICCV} and AGW \cite{ye2020deep}, both using the traditional L$_2$ distance metric in feature similarity evaluation. Table.~\ref{table:generalization} shows experimental results. As Table.~\ref{table:generalization} shows, either SGR or MNNR improves the re-ID performance by large margins when it is incorporated into the AlignGAN and AGW methods. In addition, further improvements are observed when both SGR and MNNR are incorporated. These results are well aligned with the experimental results observed in the Ablation Studies. 

Specifically, AlignGAN* + SGR achieves a mAP of 51.30\% and a rank-1 accuracy of 52.56\% which are higher than the AlignGAN by 16.74\% and 4.64\%, respectively. AlignGAN* + MNNR achieves a mAP of 50.26\% and a rank-1 accuracy of 52.33\% which also outperforms AlignGAN significantly. Similar improvement can be observed for AGW as well. All these experimental results demonstrate the superiority of our proposed Similarity Inference Metric (with SGR and MNNR) that can generalize across different cross-modality re-ID metrics with significant and consistent performance improvements but little extra training.

\begin{table}[t]
\begin{center}
\begin{tabular}{l|cc}
    \hline\hline
    Methods  & mAP & Rank-1 \\
    \hline
    AlignGAN*\cite{Wang_2019_ICCV}            &34.56 & 47.92    \\ 
    AlignGAN* + SGR      & 51.30 & 52.56   \\
    AlignGAN* + MNNR     & 50.26 & 52.33    \\
    AlignGAN* + SIM (SGR+MNNR)  &\textbf{54.45}  & \textbf{52.70} \\ 
    \hline
    AGW  ~\cite{ye2020deep}  &40.03 & 50.87    \\ 
    AGW + SGR      & 55.89 & 52.70   \\
    AGW + MNNR     & 51.40 & 52.93    \\
    AGW + SIM (SGR+MNNR)      &\textbf{57.47}  & \textbf{53.75}    \\ 
    \hline\hline
\end{tabular}
\end{center}
\vspace{-0.4cm}
\caption{Generalization Analysis of Similarity Inference Metric on SYSU-MM01 dataset with other approach as baseline, \textit{i.e.} AlignGAN. AlignGAN* denotes our re-implemented version.}
\label{table:generalization}
\vspace{-0.4cm}
\end{table}



\section{Conclusion}
This paper presents an innovative Similarity Inference Metric (SIM) for RGB-Infrared person re-identification. We introduce similarity graph reasoning and mutual nearest-neighbor reasoning to infer inter-modality sample similarities by exploiting reliable intra-modality sample similarity. The two types of reasoning can generalize over different cross-modality person re-ID metrics with significant performance improvements but little extra training. Experiments demonstrate the effectiveness as well as the generalization of our method for improving re-ID performance. We expect that the proposed SIM will inspire new insights for better cross-modality person re-ID in the near future. 

\section*{Acknowledgement}
This work was supported in part by National Natural Science Foundation of China (61902009) and Shenzhen Research Project (201806080921419290).


\bibliographystyle{named}
\bibliography{ijcai20}

\end{document}